\documentclass[10pt,twocolumn,letterpaper]{article}
\PassOptionsToPackage{pagenumbers}{cvpr}
\author{Yiwei Liu\\
School of Science and Engineering\\
The Chinese University of Hong Kong, Shenzhen\\
{\tt\small yiweiliu1@link.cuhk.edu.cn}}
\newcommand{\paperpdfauthor}{Yiwei Liu}

\usepackage{cvpr}
\usepackage{fontspec}
\usepackage{array}
\usepackage[nopatch=footnote]{microtype}
\usepackage{xurl}
\definecolor{cvprblue}{rgb}{0.21,0.49,0.74}
\usepackage[pagebackref,breaklinks,colorlinks,allcolors=cvprblue]{hyperref}
\hypersetup{pdfauthor={\paperpdfauthor},pdfsubject={},pdfkeywords={}}
\hypersetup{pdftitle={Beyond Readability: Evaluating Task Information Recoverability}}
\title{Beyond Readability: Evaluating\\Task Information Recoverability}
\newcommand{\authorfield}{\textsc{AUTHOR}}
\begin{document}
\maketitle

\begin{abstract}
Direct visual readability and task-information recoverability are different quantities. Failure to decode a target from a fixed observation need not eliminate access to that target through another recovery route. We develop an evaluation perspective that makes the observation, query, target, and available knowledge explicit and measures the overlap between routes' success sets. For information available on the original visible surface under suitable imaging conditions, direct optical recovery reads the target from the image, optionally after restoration; entity-linked recovery uses residual visual evidence to identify the depicted entity and accesses its target through an entity--attribute relation in a specified knowledge resource. Such access can draw on stored knowledge or an external source. A controlled book-cover study instantiates external access with a fixed title--author catalog, comparing optical author recovery with visual title resolution and deterministic lookup under resolution degradation. Entity-linked successes persist across the tested vision--language models, revealing information access beyond the tested direct visual frontier despite substantial differences in absolute performance. A substantial optical-only region remains. These complementary outcomes show why visual degradation should be evaluated through the task information accessible along specified routes and knowledge resources, alongside direct readability.
\end{abstract}

\section{Introduction}
\label{sec:intro}

Direct visual readability and task-information recoverability answer different questions. Readability concerns decoding a target from its visible representation; recoverability concerns access to that target from an observation through a specified route. A fixed image may frustrate direct optical decoding while retaining cues that identify the depicted entity. That identity can index an entity--attribute relation in stored knowledge or an external resource, making the requested target accessible without decoding its surface representation. An evaluation of perception therefore needs to compare complete recovery routes on the same observation as well as measure their individual accuracy.

Human perception motivates this distinction. Sensory evidence is interpreted through experience, context, and knowledge: an incomplete image can still support a meaningful account of the object that produced it~\citep{gregory1980perceptions,kersten2004object}. Theories of object recognition explain how visual structure supports identity across changes in appearance, while studies of the ventral visual pathway characterize representations that support this stability~\citep{biederman1987recognition,dicarlo2012object,grillspector2014architecture}. People can also extract category or scene meaning from very brief presentations~\citep{thorpe1996speed,potter2014meaning}. These findings motivate a task-centered evaluation question: when fine detail becomes difficult to decode, what useful information remains accessible from the same observation?

Context makes this question especially important. Experiments on object identification show that surrounding scenes and the consistency of object--scene relations influence recognition~\citep{palmer1975context,biederman1982scene,davenport2004scene}. Context constrains plausible interpretations, allowing an object's appearance to be understood in relation to its environment and learned associations~\citep{bar2004context,oliva2007context}. Top-down influences can facilitate visual recognition, including through information available before detailed identification is complete~\citep{bar2006topdown,gilbert2013topdown}. Experience also guides the allocation of attention: learned spatial regularities, scene knowledge, and task goals shape which evidence is selected~\citep{chun1998contextual,henderson2003gaze,wolfe2017attention}. Together, these perspectives suggest that evaluating perception requires attention to the relationship between available visual cues and the knowledge that makes those cues useful.

Predictive and inferential accounts make this relationship explicit. Predictive coding describes interactions between sensory signals and predictions about their causes~\citep{rao1999predictive,friston2005theory,clark2013whatever}, while work on associative prediction and perceptual expectation examines how prior experience shapes interpretation and decisions~\citep{bar2007proactive,summerfield2014expectation,delange2018expectations}. Findings on expectation-dependent visual representations and recurrent object recognition further emphasize that recognition involves more than a single pass through the incoming signal~\citep{kok2012less,kar2019recurrent}. For evaluation, the resulting motivation is straightforward: the detail that a particular decoder can extract and the information that a system can access using that observation are distinct quantities.

\begin{figure*}[t]
\centering
\includegraphics[width=\linewidth]{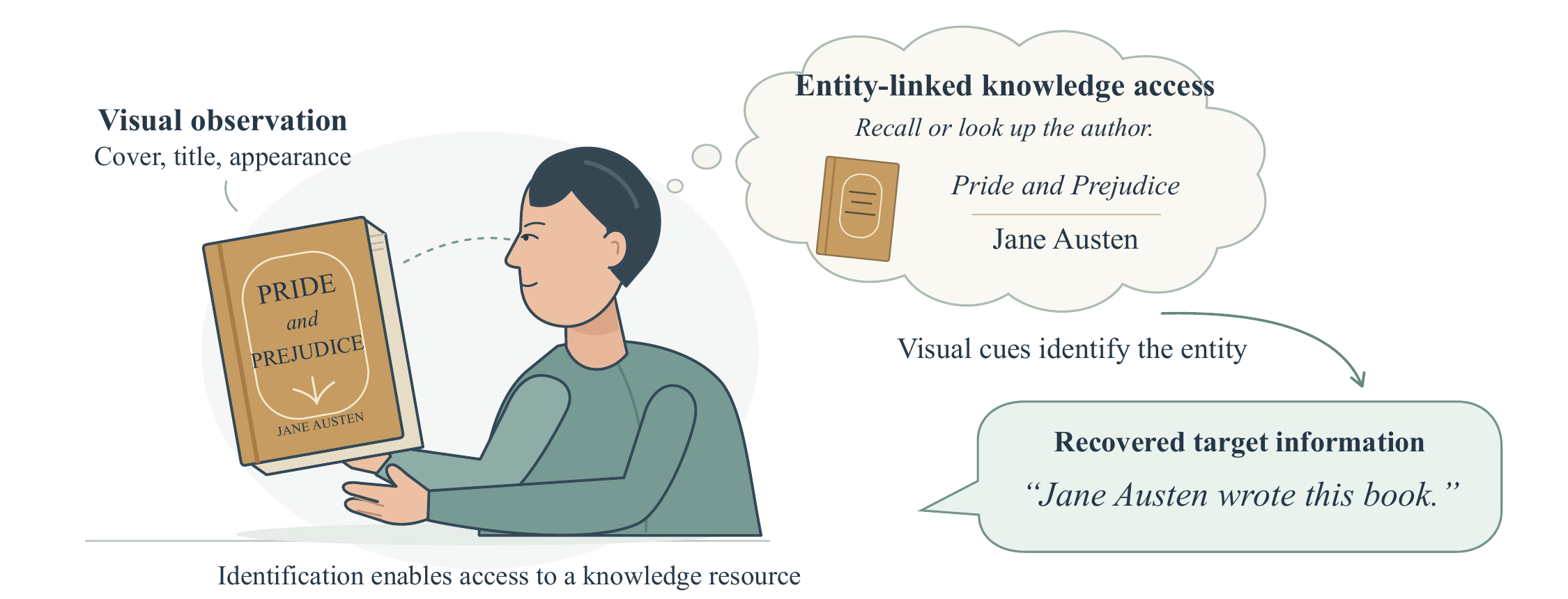}
\caption{Visual cues support book identification, which in turn enables access to an associated author through existing knowledge or an external resource. The illustration motivates evaluation of the information accessible from a fixed observation through a complete recovery route.}
\label{fig:human_perception}
\end{figure*}

Semantic memory provides a further connection between recognizing an object and knowing something about it. Research on semantic representation describes how perceptual inputs relate to stored knowledge of entities, attributes, and relationships~\citep{patterson2007semantic,binder2009semantic,binder2011semantic}. Accounts of semantic cognition and grounded cognition connect this knowledge to perceptual experience and its use in context~\citep{lambonralph2017semantic,barsalou2008grounded}. A book illustrates the distinction: its printed author name may be difficult to transcribe, yet its title, artwork, or layout may identify the book. Its identity can then provide access to the author through existing knowledge or an external lookup, as illustrated in Figure~\ref{fig:human_perception}.

Modern vision--language systems offer a practical setting for studying such access. Large-scale image--text learning connects visual representations to language, and multimodal models combine visual inputs with pretrained language capabilities~\citep{radford2021clip,jia2021align,alayrac2022flamingo}. Visual entity recognition and information-seeking question answering already use images to address information about particular entities~\citep{hu2023oven,chen2023infoseek,mensink2023encyclopedic}.

We operationalize this distinction through two ways of accessing the same target. \emph{Direct optical recovery} decodes the target from its visual representation, optionally after restoration. \emph{Entity-linked knowledge recovery} uses residual visual evidence to identify the entity and then accesses the target through a specified entity--attribute resource. We restrict attention to information available on the original visible surface under suitable imaging conditions. In our controlled demonstration, a model resolves a book title from the image and a fixed external title--author catalog supplies the author. This choice makes knowledge access deterministic after entity linking while controlling the source and retrieval variability of open search. A \emph{recoverability gap} appears when the entity-linked route accesses a target that the tested direct decoders fail to recover from the same observation. Joint outcomes expose this gap and the complementary optical-only region, which marginal accuracy alone leaves unresolved.

We contribute an evaluation framework that treats direct decoding and entity-linked knowledge access as complete routes from a fixed observation to the same target. A controlled external catalog makes the entity--attribute relation explicit and auditable, so success requires both the correct identity and the correct target. A Strong Optical Frontier summarizes success across the tested direct decoders; joint outcome counts and conditional rescue rates then measure which observations support either route or both.

We demonstrate this perspective with a controlled evaluation of 5,000 books across seven degradation levels, providing 35,000 observations and using a fixed catalog for external knowledge access. The primary Qwen3.7-Plus~\citep{alibaba2026models} evaluation shows entity-linked recovery within the optical-failure region alongside a substantial optical-only region. The optical frontier has higher overall accuracy, but the entity-linked route exceeds it at the strongest degradation levels. Replacing the title-resolution model with Qwen2.5-VL-32B-Instruct~\citep{bai2025qwen25vl} or InternVL3.5-14B-Instruct~\citep{wang2025internvl35} preserves a nontrivial entity-linked-only region under the same frozen protocol, despite substantial variation in absolute performance. Complementary recoverability is thus observed across all three tested VLMs: direct visual decoding and visual identification followed by catalog access expose partially distinct regions of task information access.

\section{Related Work}
\label{sec:related}

\subsection{Restoration and text recovery}

General super-resolution methods provide plausible routes to improve the input of a text decoder. HAT combines channel and window-based self-attention with overlapping cross-attention~\citep{chen2023hat}, while Real-ESRGAN learns blind restoration using synthetic degradations~\citep{wang2021realesrgan}. We evaluate their restored images through the same OCR stack as the native input. Image fidelity itself is not the scored target.

Text-specific restoration is a relevant alternative to these general routes. TextZoom and Scene Text Telescope study reconstruction for reading scene text~\citep{wang2020textzoom,chen2021stt}; more recently, TIGER separates glyph restoration from full-image enhancement~\citep{luo2026tiger}. These approaches emphasize the importance of evaluating restoration through the content that a downstream decoder actually recovers. Our evaluation compares the selected general restoration routes using a common target-recovery criterion.

\subsection{Visual entity recognition and structured knowledge access}

Knowledge-based visual question answering connects image understanding with information beyond directly visible content. OK-VQA and A-OKVQA study questions requiring external or world knowledge~\citep{marino2019okvqa,schwenk2022aokvqa}, while InfoSeek and Encyclopedic VQA emphasize detailed information about fine-grained entities~\citep{chen2023infoseek,mensink2023encyclopedic}. Entity recognition provides a link between visual evidence and such information: OVEN maps an image and query to a Wikipedia entity, and WikiCLIP develops a contrastive approach to open-domain visual entity recognition~\citep{hu2023oven,ning2026wikiclip}.

Our study uses this recognition-to-knowledge connection to evaluate recovery of a surface-available target under degradation. Entity identity can index an attribute in existing knowledge or in an external resource; the controlled experiment instantiates the latter with exact lookup in a fixed title--author catalog. This choice holds coverage, title ambiguity, and answer mapping constant rather than introducing variation from open-web retrieval. Residual visual evidence supports book-title resolution, and successful linking provides access to the author. Comparing this complete route with direct optical decoding on the same observations reveals where entity-linked knowledge recovers information that the tested optical routes fail to read.

\section{Recovery Framework}
\label{sec:method}

We evaluate recovery of task-relevant information from a single fixed RGB observation and a natural query. The query requests a textual or text-serializable target with its own readable or decodable representation on the depicted entity's original visible surface under suitable imaging conditions. A complete recovery route may decode the target directly or use the observation to identify the entity and access its target through a specified knowledge resource. The output is the requested information or an abstention.

\subsection{Direct Optical Recovery}

Direct optical recovery obtains the target by decoding its surface representation in the native image or a resized or restored version, using learned or conventional processing. Success is measured by recovery of that content, independently of the visual quality of an intermediate image.

\subsection{Entity-linked Knowledge Recovery}

The entity-linked semantic route uses residual visual evidence to identify the depicted entity and then accesses its associated target through an entity--attribute relation. Partial text, appearance, and layout can support identification even when direct target decoding fails. The knowledge resource may be internal to the recognizer or external to it; in the controlled study, a fixed catalog serves as the external resource. The model resolves a book title from the degraded image, and an exact catalog match supplies the author. Entity ambiguity, incorrect linkage, insufficient evidence, and abstention are possible outcomes of this complete route.

The scientific distinction is how the requested information is accessed: direct decoding of its surface representation or visual identification followed by access through an entity-to-target relation. Joint outcomes measure where these routes succeed together and where either route alone recovers the target. This comparison captures complementary recoverability from the same observation.

\section{Controlled Evaluation}
\label{sec:setup}

The controlled evaluation measures author recovery from 5,000 book covers across seven resolution levels, yielding 35,000 observations with entity-linked scoring. The observation set and title--author catalog are fixed throughout; the supplement specifies construction, inference settings, hardware, and scoring in detail.

\subsection{Evaluation design}
\label{sec:evaluation}

Native, HAT x4, and Real-ESRGAN x4 start from the same RGB observation and use the same PaddleOCR stack. OCR processes each full route image at relative scales 1 and 2, with LANCZOS resizing and deterministic transcript concatenation. The target matcher tests normalized substring containment, ignoring case, spacing, and punctuation. For evaluation, we define the Strong Optical Frontier as the union of successes across the tested optical routes. Intended-set rescue divides semantic successes by all optical-frontier failures; no-response cases are reported separately.

\subsection{Observation set and degradation}
\label{sec:controlled_setup}

The target is the \authorfield{} printed on a book cover. OCR-VQA metadata~\citep{mishra2019ocr} are pooled into a catalog of 184,809 title-unique entries, excluding ambiguous normalized titles without author-based disambiguation. Deterministic ordering selects 8,000 candidates; clean-cover OCR identifies 5,785 eligible books whose normalized author occurs in the transcript. The evaluation freezes 5,000 unique books before degradation, with no model training or adaptation.

Each book yields seven observations at scales 1.0, 0.75, 0.5, 0.35, 0.25, 0.18, and 0.125. BOX/area downsampling is followed by aspect-preserving LANCZOS resizing and centering on a black 512-by-512 RGB canvas, without JPEG recompression. Scale 1.0 includes canvas resizing and differs from the clean-screen image. All routes share the 35,000 observations, with equal weight across levels.

\subsection{Entity-linked semantic recovery}

Qwen3.7-Plus through the official Alibaba Cloud service receives only the degraded image and query and returns a title hypothesis, visible anchors, and abstention. Exact title linking applies Unicode NFKC, case folding, and removal of non-alphanumeric characters; a unique match in the fixed external catalog supplies the author. Thus the model identifies the book from the image while the catalog provides the entity--attribute mapping. Success requires both the correct book and author, so a wrong book sharing the author fails. Evaluation entities are intentionally in the catalog, and incomplete or variant titles can remain unresolved.

Semantic evaluation yields 34,934 valid responses and 66 infrastructure/API failures. Incorrect valid responses are retained without retries.

To assess robustness to the semantic model, we also evaluate Qwen2.5-VL-32B-Instruct and InternVL3.5-14B-Instruct on the same 35,000 observations. Both use image-based title resolution followed by the same deterministic catalog lookup and entity-linked scoring.

\section{Results}
\label{sec:results}
\label{sec:controlled_results}

\begin{figure*}[t]
\centering
\includegraphics[width=\linewidth]{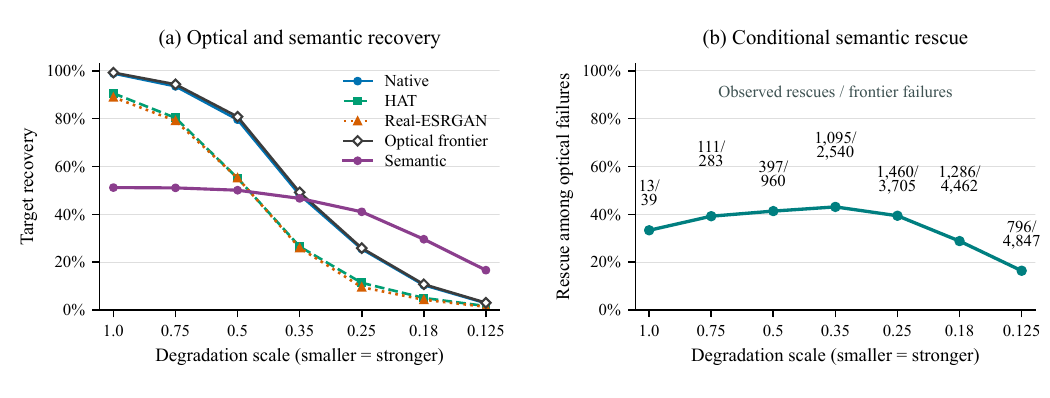}
\caption{Controlled recovery under fixed degradation, using Qwen3.7-Plus for the semantic route. Left: all routes on the full intended set of 5,000 observations per level. Right: semantic success conditioned on failure of the Strong Optical Frontier, with the number of observed rescues over all frontier failures shown at each point. Smaller scales mean stronger degradation; points denote the seven tested levels.}
\label{fig:curves}
\end{figure*}

\paragraph{Recovery regions.}
Table~\ref{tab:overall} gives marginal target recovery on the controlled observation set for the primary Qwen3.7-Plus evaluation. The optical frontier improves only slightly on Native, since restoration adds relatively few successful cases. Semantic recovery has lower overall accuracy, with entity-correct and catalog-derived author-correct counts coinciding. The joint outcomes reveal how this marginal ordering relates to recovery on individual observations.

\begin{table}[t]
\caption{Controlled evaluation: overall target recovery. A semantic success requires the correct entity and author.}
\label{tab:overall}
\centering
\small
\includegraphics{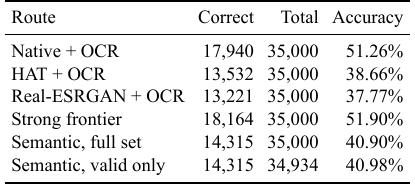}
\end{table}

The joint outcomes in Table~\ref{tab:overlap} expose the distinction. Visual identification followed by catalog access recovers 5,158 of the 16,836 observations where all tested optical routes fail, giving 30.64\% intended-set rescue. A substantial optical-only region remains as well, so neither observed success set contains the other. This is evidence of complementary task-information recovery despite the entity-linked route's lower marginal accuracy.

\begin{table}[t]
\caption{Controlled evaluation: joint target-recovery outcomes. The first two columns form the two-by-two table for the 34,934 observations with valid semantic responses.}
\label{tab:overlap}
\centering
\small
\includegraphics{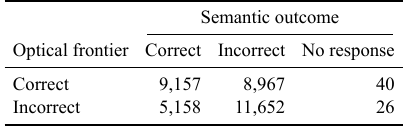}
\end{table}

\begin{figure*}[t]
\centering
\includegraphics[width=\linewidth]{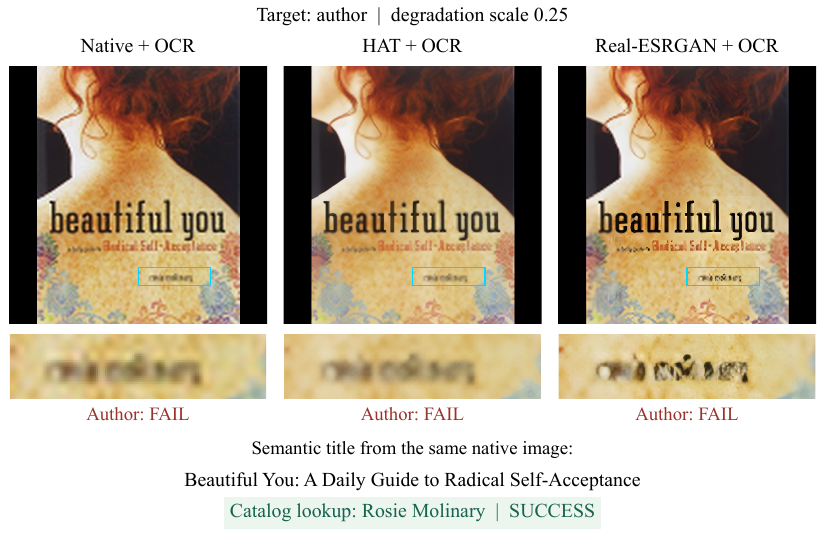}
\caption{Controlled evaluation: author recovery for \emph{Beautiful You}. All three full archived route images are shown, with matched author-region enlargements. Native, HAT, and Real-ESRGAN OCR each fail the frozen author matcher. Semantic title resolution uses the same native degraded image; exact linking to the frozen catalog supplies Rosie Molinary, with both entity and author correct.}
\label{fig:controlled_qualitative}
\end{figure*}

\paragraph{Effect of degradation.}
Figure~\ref{fig:curves} and Table~\ref{tab:levels} show how the observed regions change with resolution. Optical recovery falls sharply as scale decreases, while semantic recovery initially declines more slowly and exceeds the frontier at the three smallest scales. Both routes fail frequently at the strongest degradation, indicating that semantic access also depends on the surviving evidence.

\begin{table*}[t]
\caption{Controlled evaluation by degradation level, with 5,000 observations per row. Accuracies use the full intended set. Rescue is the semantic-success count divided by all frontier failures at that level.}
\label{tab:levels}
\centering
\small
\includegraphics{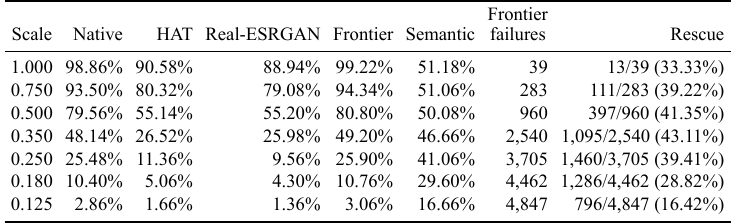}
\end{table*}

Conditional rescue peaks at scale 0.35, whereas the largest absolute rescue count occurs at 0.25. The distinction reflects the changing size and composition of the optical-failure subset, which contains only 39 observations at scale 1.0. Declining rescue at the smallest scales is consistent with increasingly ambiguous identity evidence.

\paragraph{Semantic failure modes.}
The outcome taxonomy contains 14,315 successes, 19,243 unresolved titles/entities, 707 wrong entities, 669 abstentions, and 66 infrastructure/API failures. Unresolved titles dominate scientific errors even at scale 1.0. Unresolved outcomes may arise from incomplete title recovery or unsuccessful entity linking. Abstention rises sharply at the strongest degradation, adding a distinct failure mode as the image becomes less informative.

\paragraph{Qualitative evidence.}
Figure~\ref{fig:controlled_qualitative} illustrates one frozen rescue at scale 0.25. The three optical transcripts retain title fragments but miss the author. From the same degraded input, the entity-linked route resolves the full title, enabling exact catalog lookup of the correct author. This case illustrates how surviving visual evidence can identify an entity and thereby index the target in an external resource when direct decoding fails.

\paragraph{Semantic-model robustness.}
Table~\ref{tab:semantic_robustness} tests whether semantic-only recovery persists when the title-resolution model changes. Absolute author recovery and rescue rates vary substantially, yet every tested VLM recovers the target on a nontrivial subset of the same optical-frontier failures.

\begin{table}[t]
\caption{Semantic-model robustness on the same 35,000 observations and frozen optical frontier. AUTHOR correctness requires the correct book and catalog-derived author. Rescue uses all 16,836 frontier failures.}
\label{tab:semantic_robustness}
\centering
\small
\includegraphics{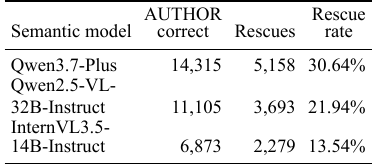}
\end{table}

\section{Discussion}
\label{sec:discussion}

\subsection{Interpreting recovery routes}

The recoverability gap changes how failure under visual degradation should be interpreted. Failure of the tested optical routes identifies a boundary of direct decoding; it leaves open whether the same target is accessible through another route from the same observation. The controlled study makes this distinction observable: an entity-linked-only region persists across all three tested VLMs, with its size depending strongly on the model, while a substantial optical-only region remains. These complementary success sets make route overlap essential to interpreting information access, even when one route has higher marginal accuracy.

The relevant unit of evaluation is the complete recovery route, with its observation, query, and permitted knowledge resource. Here, residual visual evidence supports book identification, and a fixed external catalog supplies the author through an entity--attribute relation. The book-cover evaluation thus gives a controlled instance of the broader principle that readability alone cannot determine task-information recoverability. Evaluating degraded observations through the information still accessible along specified routes avoids equating poorer visual detail with automatic loss of the requested information.

\subsection{Open-web knowledge access}

Open-web search services such as Jina Search and DuckDuckGo could provide broader and more current entity-linked evidence than a fixed catalog, extending access when a target relation is absent from a curated resource. Their use also makes controlled evaluation harder: queries generated from the image, result ranking, page availability and content, and a model's selection and interpretation of retrieved evidence can all affect the final answer. We therefore use a fixed catalog as the external knowledge source in the present study, making entity coverage and title--author mapping explicit and lookup deterministic. A useful next step is to evaluate open-web retrieval under the same fixed-observation protocol, varying the search service, query strategy, and evidence selection while recording the retrieved sources and their effect on information recovery.

\subsection{Scope of surface-grounded targets}

Our evaluation focuses on targets with a decodable representation on the original visible surface under suitable imaging conditions. A plain cup, for example, can be recognized from its shape even though the word ``cup'' is not printed on it; asking what the object is poses an object-recognition question without a corresponding surface field to read. Such questions are important, but they require a different reference route and controls for comparing visual recognition with knowledge access. We restrict the present study to surface-grounded textual or text-serializable targets so that direct decoding and entity-linked access can be evaluated against the same target from the same fixed observation.

\section{Conclusion}
\label{sec:conclusion}

Readability is not recoverability: failure of direct optical decoding can coexist with task information accessible through another route from the same fixed observation. Visual identification can index an entity--attribute relation in a knowledge resource, enabling access to a target whose surface representation is difficult to decode. Our controlled book-cover study instantiates this route through title resolution and external lookup in a fixed catalog, alongside direct optical author recovery. Entity-linked-only successes persist across the tested VLMs, while a substantial optical-only region remains. The broader evaluation lesson is to measure task information through complete routes and their specified knowledge resources rather than equate visual degradation with its loss.

\subsection*{AI Use Disclosure}

As English is not the authors' native language, AI-assisted tools were used to improve language clarity, grammar, and overall readability of the manuscript. The use of AI was limited to writing assistance and does not extend to the development of the research ideas, experimental design, data analysis, or scientific conclusions. All AI-assisted revisions were reviewed and verified by the authors, who are fully responsible for the final manuscript.

{\small
\bibliographystyle{ieeenat_fullname}
\bibliography{references}
}
\end{document}


\maketitle

\section{Controlled-set construction}
\label{sec:controlled_construction}

\subsection{Catalog and candidate selection}

The controlled task recovers the author printed on a book cover using metadata from OCR-VQA. Metadata from the train, validation, and test splits are pooled for evaluation; no model is trained on these splits. For repeated image identifiers, the representative is the first record in lexicographic split order and then integer row order. Records with an empty title, author, or normalized title are removed.

Title normalization applies Unicode NFKC, case folding, and retention of alphanumeric characters. Every normalized title associated with multiple book identifiers is excluded, with no author-based disambiguation. The resulting catalog contains 184,809 title-unique entries and maps each retained title to its book identity and author. It serves as the fixed external knowledge resource: once a title is linked, access to the author is deterministic rather than dependent on changing search results or source ranking.

Candidate selection uses the seed \nolinkurl{ocr-vqa-controlled-v1-20260909}. Books are ordered by the SHA-256 digest of this UTF-8 seed, followed by a null separator and the book identifier. The first 8,000 records form the candidate pool. The same ordering selects the final books after the clean-cover eligibility screen.

\subsection{Clean-cover eligibility}

Eligibility requires the normalized catalog author to occur as a substring of the normalized OCR transcript from the original clean cover. The screen uses the controlled experiment's OCR configuration described in Section~\ref{sec:ocr_config}, including full-image processing at relative scales 1 and 2. Of the 8,000 candidates, 5,785 pass this screen; the first 5,000 eligible unique books in the deterministic order form the fixed evaluation set.

This screen establishes that the requested attribute is recoverable from the original surface under the clean-image condition. Membership is fixed before degraded-image evaluation.

\section{Degradation and optical recovery}
\label{sec:optical_repro}

\subsection{Resolution degradation}

Each selected cover is decoded as RGB and processed independently at scales 1.0, 0.75, 0.5, 0.35, 0.25, 0.18, and 0.125. At each scale, the original width and height are multiplied by the scale and rounded to the nearest integer using Python's rounding convention, with each dimension clamped to at least one pixel. BOX/area downsampling produces the reduced image.

The reduced image is then resized with LANCZOS interpolation to fit a 512-by-512 canvas while preserving its aspect ratio. The fit factor is the smaller of the canvas-width and canvas-height ratios; the resized dimensions use the same rounding and minimum-pixel rule. The image is centered on a black RGB canvas using integer offsets obtained by floor division of the remaining width and height by two. Outputs are saved losslessly as PNG, with no JPEG recompression. Scale 1.0 retains the canvas-resizing step and therefore differs from the image used for eligibility screening.

All recovery routes receive the same resulting observation for a given book and scale. The design contains 35,000 observations, with 5,000 at each level; overall observation-level rates weight the seven levels equally. Each image is evaluated independently, including across different levels of the same book.

\subsection{Restoration and hardware}

The optical routes process the native observation directly, apply HAT x4, or apply Real-ESRGAN x4 before OCR. HAT uses the pretrained Real HAT GAN x4 checkpoint, and Real-ESRGAN uses the pretrained x4plus checkpoint. Both use tile size 256; padding is 32 pixels for HAT and 10 pixels for Real-ESRGAN. Restoration and OCR models are used without training or adaptation to the evaluation set.

The optical restoration experiments use a workstation with five NVIDIA GeForce RTX 2080 Ti GPUs, each with 11,264 MiB of memory, and NVIDIA driver 570.124.04. Controlled OCR evaluation runs on the CPU.

\subsection{OCR configuration}
\label{sec:ocr_config}

The optical routes use PaddleOCR 3.7.0 with the PP-OCRv6 medium detector and recognizer, configured for Chinese and English text through the Chinese language setting, with PaddlePaddle 3.3.1. Table~\ref{tab:ocr_settings} gives the OCR processing settings shared by all routes.

\begin{table}[t]
\caption{OCR configuration for native and restored route images.}
\label{tab:ocr_settings}
\centering
\small
\includegraphics{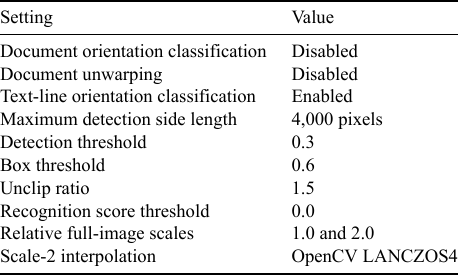}
\end{table}

OCR processes each complete route image first at its native route resolution and then at twice that resolution, using the same preprocessing and matching policy for every observation. Recognized lines are concatenated in scale order and, within each scale, the model's reading order.

For optical target matching, normalization applies Unicode NFKC and uppercase conversion, removes whitespace, and retains alphanumeric characters and code points U+3400--U+9FFF. A nonempty normalized target must occur as a substring of the normalized OCR transcript.

\section{Controlled semantic recovery}
\label{sec:controlled_semantic}

\subsection{Primary model and service settings}

The controlled semantic route uses Qwen3.7-Plus through the official Alibaba Cloud Bailian/DashScope service in the Beijing region. Requests use temperature zero, reasoning disabled, and a maximum output budget of 1,800 tokens. The evaluation was completed in September 2026.

Each request contains only the degraded image and query, asking for a title hypothesis, visible anchors, and an abstention decision. The model uses residual visual evidence to resolve the book identity; subsequent deterministic lookup in the fixed external title--author catalog supplies the author. The complete route therefore combines visual entity identification with controlled knowledge access.

\subsection{Title linking and target scoring}

The predicted title is normalized using Unicode NFKC, case folding, and retention of alphanumeric characters, matching the catalog construction rule. An exact normalized-title match to a unique record links the observation to that book and supplies its author. Incomplete subtitles and alternative title forms can therefore remain unresolved even when they resemble a catalog title.

For a valid semantic response, success requires a non-abstained prediction linked to the correct book and yielding the correct author. A link to a different book is a wrong-entity outcome even if that book shares the target author. The semantic outcome categories are success, unresolved title or entity, wrong entity, abstention, and infrastructure/API failure. Incorrect valid responses are retained without retries; provider or transport failures with no valid response are counted separately.

\section{Local VLM robustness}
\label{sec:local_vlm}

The local evaluations use Qwen2.5-VL-32B-Instruct at revision {\urlstyle{same}\nolinkurl{7cfb30d71a1f4f49a57592323337a4a4727301da}} and InternVL3.5-14B-Instruct at revision {\urlstyle{same}\nolinkurl{72c82460a2c05a3bc2b47be230c44edc88e86ed4}}, with BF16 inference. Both use deterministic decoding (\texttt{do\_sample=false}, \texttt{temperature=0}, \texttt{max\_new\_tokens=1800}), the same semantic prompt, title-linking rule, and scoring protocol as the primary evaluation. We retain first-pass results only and perform no targeted retry. The local VLM evaluations used four NVIDIA A100-SXM4-40GB GPUs. Table~\ref{tab:local_validity} reports the first-pass response accounting for both local models.

\begin{table}[!htbp]
\caption{First-pass response accounting for the two local VLMs, with 35,000 observations per model.}
\label{tab:local_validity}
\centering
\small
\includegraphics{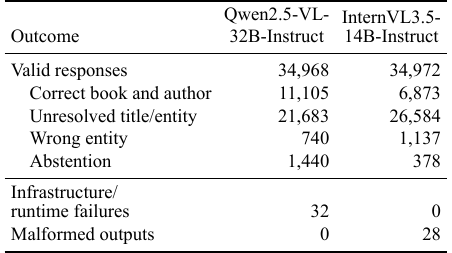}
\end{table}